\documentclass{article}

\usepackage{arxiv}

\usepackage[utf8]{inputenc} 
\usepackage[T1]{fontenc}    
\usepackage{hyperref}       
\usepackage{url}            
\usepackage{booktabs}       
\usepackage{amsfonts}       
\usepackage{nicefrac}       
\usepackage{microtype}      
\usepackage{lipsum}		
\usepackage{graphicx}
\usepackage{natbib}
\usepackage{doi}
\usepackage{amsmath}
\usepackage{algorithm}
\usepackage{algpseudocode}

\title{Refined Iterated Pareto Greedy for Energy-aware Hybrid Flowshop Scheduling with Blocking Constraints}

\author{ {\includegraphics[scale=0.06]{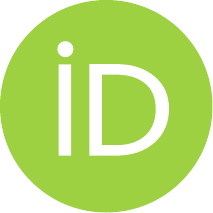}\hspace{1mm} Ahmed Missaoui}\thanks{Use footnote for providing further
		information about author (webpage, alternative
		address)---\emph{not} for acknowledging funding agencies.} \\
	Department of Computer Science\\
    University College Cork\\
    Cork, T12 XF62, Ireland\\
	\texttt{amissaoui@ucc.ie} \\
	\And
	{\includegraphics[scale=0.06]{orcid.pdf}\hspace{1mm}Cemalettin Ozturk} \\
	Process, Energy and Transport Engineering \\
    Munster Technological University \\
	Bishopstown, Cork, T12 P928, Ireland \\
	\texttt{cemalettin.ozturk@mtu.ie} \\
    \And
	{\includegraphics[scale=0.06]{orcid.pdf}\hspace{1mm}Barry O'Sullivan} \\
	Department of Computer Science\\
    University College Cork\\
    Cork, T12 XF62, Ireland\\
	\texttt{b.osullivan@cs.ucc.ie} \\
}

\renewcommand{\shorttitle}{\textit{arXiv} Template}

\hypersetup{
pdftitle={A template for the arxiv style},
pdfsubject={q-bio.NC, q-bio.QM},
pdfauthor={David S.~Hippocampus, Elias D.~Striatum},
pdfkeywords={First keyword, Second keyword, More},
}

\begin{document}
\maketitle

\begin{abstract}
	The scarcity of non-renewable energy sources, geopolitical problems in its supply, increasing prices, and the impact of climate change, force the global economy to develop more energy-efficient solutions for their operations. The Manufacturing sector is not excluded from this challenge as one of the largest consumers of energy. Energy-efficient scheduling is a method that attracts manufacturing companies to reduce their consumption as it can be quickly deployed and can show impact immediately. In this study, the hybrid flow shop scheduling problem with blocking constraint (BHFS) is investigated in which we seek to minimize the latest completion time (i.e.~makespan) and overall energy consumption, a typical manufacturing setting across many industries from automotive to pharmaceutical. 
Energy consumption and the latest completion time of customer orders are usually conflicting objectives.
Therefore, we first formulate the problem as a novel multi-objective mixed integer programming (MIP) model and propose an augmented epsilon-constraint method for finding the Pareto-optimal solutions. Also, an effective multi-objective metaheuristic algorithm,  \textquotedblleft Refined Iterated Pareto Greedy (RIPG)\textquotedblright, is developed to solve large instances in reasonable time. Our proposed methods are benchmarked using small, medium, and large-size instances to evaluate their efficiency. Two well-known algorithms are adopted for comparing our novel approaches. The computational results show the effectiveness of our method.
\end{abstract}

\keywords{Hybrid flow shop
\and Blocking constraint \and Energy consumption \and Mixed integer linear programming \and Iterated Pareto greedy}

\section{Introduction}\label{Intr}
Growth in the global economy, and technological developments on new products and services provided to customers,
 are constantly increasing the energy demand for the manufacturing industry. Industry consumes almost 30\% of global energy according to \cite{meng2019milp} which is mostly produced from non-renewable resources. Therefore, scarcity and geopolitical risks to energy supply as well as high emission levels impose pressure on the manufacturing industry to reduce energy consumption.

The majority of the effort on reducing manufacturing energy
consumption has been focusing on more energy-efficient machines, equipment and processes, but as stated by \cite{JMS05}, more than 85\% of the energy in mass production systems is consumed by non-production related functions. This finding indicates that there is a significant opportunity to reduce manufacturing energy consumption by focusing on system-level improvements rather than individual components. Improving manufacturing planning and scheduling by taking into account energy efficiency is one such system-level improvements that has
 an immediate impact and hence attracts both practitioners and researchers(~\cite{framinan2010architecture,lei2017novel,liu2020mixed}). As a result, manufacturing companies are gradually increasing energy awareness in their planning and scheduling decisions to minimize their consumption without sacrificing their quality of service (\cite{shrouf2014optimizing}).




In a recent study, ~\cite{missaoui2024energy} classified different energy-efficient scheduling applications. This taxonomy shows that short and medium-term strategies are utilized across various industries from 
metal~\cite{gomes2021multi} to food 
production~\cite{yaurima2018hybrid}.
Energy efficiency in discrete-manufacturing systems, such as flexible job shop manufacturing, is reviewed by~\cite{JMSone} and~\cite{JMSthree} in which strategies for reducing energy consumption and optimizing the use of resources are discussed both from academic and industrial perspectives. The growing interest in energy-efficient scheduling reflects its critical importance in sustainability and cost reduction.
Therefore, recent studies in the literature aim to approach this multi-dimensional problem as a multi-objective optimization problem~\cite{ASCOne}.

In practical settings, manufacturing is realized by a set of production stages. In the basic case, each stage includes one resource able to perform a step of the production process. This simple architecture leads sometimes to the emergence of the bottleneck stage that can decrease productivity~\cite{yu2018genetic}.
To avoid this impact, industries apply an intuitive solution by adding parallel resources. This setting is denoted as the hybrid flowshop scheduling system (HFS).
The HFS scheduling problem combines the characteristics 
of a simple flowshop problem and a parallel machines problem. Due to its flexibility and adaptability to customized production plans, HFS systems are widely employed in different industries such as steel-making \cite{guirong2017solving}, glass production \cite{wang2020energy}, and food industry \cite{yaurima2018hybrid}. Therefore, HFS is an active research area for the scientific community as well (\cite{AMMtwo,missaoui2025energy}).  


Formally, a HFS problem instance
 consists of a set of $n$ jobs that have to be handled in $g > 1$ production stages. Each stage contains $m_g$ machines where $m_g > 1$ in at least one stage. In its basic form, $n$, $g$, $m$ as well as the processing time are known in advance. The study by \cite{gupta1988two} was one of the earlier works that investigated the two stages HFS problem and demonstrated that it is NP-Hard. 

Practically, the manufacturing environment, the nature of machines, and, in some cases, the product itself, could impose some constraints that have to be respected during the scheduling process. Narrow buffer areas between production stages inevitably lead to limitations in storage. In its extreme case this restraint is well known in the literature as a blocking or no-buffer constraint. In that case, the job has to stay on the machine after the processing if there are no free machines in the next stage and the machine has to be blocked by the job. Therefore, the machine could be in three states: idle, processing, or blocking.
A detailed explication of blocking constraints in scheduling problems is given in the literature~\cite{hall1996survey}. However, some investigations of this constraint in the HFS are
available~\cite{yuan2009application,wang2011hybrid,missaoui2023energy}. 

Summing up, in this study, we address the blocking HFS scheduling problem, denoted as BHFS, focusing on minimizing the makespan and the total energy consumption (TEC), which includes the processing energy, idle time energy, and blocking energy.
The main contributions of this paper are outlined as follows:

\begin{itemize}
    \item We propose a bi-objective hybrid flowshop scheduling problem that includes blocking constraints and is formulated as novel multi-objective mixed-integer linear programming (MILP) model. The objectives are to minimize both the makespan and the overall energy consumption.

    \item We implemented the augmented epsilon constraint method to find optimal Pareto fronts via the developed MILP model.

    \item We introduce a novel scalable heuristic approach for identifying feasible Pareto fronts rapidly. We also discussed a detailed analysis of algorithm parameters for calibration.

    \item We provide an extensive number of instances to benchmark the novel MILP and metaheuristic methods against the most widely used multi-objective algorithms from the scheduling literature.
    
\end{itemize}

The remainder of this article is organized as follows, Section~\ref{Review} is dedicated to a general literature review of the problem. A mixed integer linear programming model and an illustrative example are given in the Section~\ref{MILP}. The novel solution method developed for large instances is described in Section~\ref{R-MOIG}. The experimental settings and computational results are given in Section~\ref{CompEXP}. Finally, the main conclusion and a discussion of future research directions are given in Section~\ref{Conclusion}. 

section{Literature Review}\label{Review}
In manufacturing, energy-efficient scheduling refers to the practice of using optimization methods in order to improve energy consumption in both processing and non-processing periods.
This concept appeared for the first time in \cite{boukas1990hierarchical}. In that work, a case study of HFS with power constraints was tackled in order to minimize the maximum completion time. 
Different production systems and practical energy aspects were studied further. In \cite{troncoso2004application},
authors developed a mathematical model and an evolutionary algorithm in order to minimize the 
total energy cost on a shop floor.
The HFS was investigated in \cite{liu2008mathematical} aiming to minimize the total energy consumption using a mixed integer linear programming (MILP) model and an improved genetic algorithm (GA).
In \cite{dai2013energy}, the multi-objective flexible HFS was studied to minimize the makespan and total energy consumption (TEC) using a genetic-simulated annealing algorithm, and many energy-related aspects are considered such as the speed of machines and turn off/turn on
 decisions. 
\cite{lei2017novel} presented a Teaching-Learning-Based Optimization algorithm to solve a bi-objective HFS problem with TEC and total tardiness.
\cite{wang2020energy}, proposed a MILP model to study multi-objective metaheuristics to study the energy-aware two-stage HFS. \cite{zhou2019multi} gave an  imperialist
competitive algorithm aiming to minimize makespan and TEC in the HFS production system. The same problem with fuzzy processing time was tackled in \cite{zhou2019energy} in order to minimize the TEC and total weighted delivery penalty. Recently, \cite{wang2023solving} developed a MILP model and an improved Non-dominated Sorting Genetic algorithm (NSGA-II) to investigate the energy-efficient HFS with machine speeds. In \cite{li2023improved}, a case study of HFS in sand-casting enterprises was tackled and solved using an improved cuckoo search approach.

As seen, the energy-efficient HFS has been introduced taking into account several energy aspects in the literature. Moreover, a set of different practical constraints have been investigated in energy-aware scheduling problems. The limited storage space between production stages is among the most practical constraints in manufacturing, this latter is known as blocking constraints. \cite{elmi2014scheduling} propose a simulated annealing method to study the blocking HFS (BHFS) with makespan minimization. In \cite{aqil2021two}, the BHFS is tackled using improved nature-inspired meta-heuristics to minimize the total weighted earliness tardiness. BHFS with setup times was introduced in \cite{moccellin2018heuristic} for makespan minimization through a constructive heuristic.

\cite{han2020discrete} presented a Multi-objective evolutionary approach for the blocking flowshop when the objective is the minimization of makespan and TEC. 
\cite{qin2023energy} developed a MILP model and an improved iterated greedy for the distributed HFS with blocking constraints. 

Recently, \cite{missaoui2023energy} addressed the BHFS with makespan and TEC minimization using two different metaheuristics namely multi-objective iterated greedy and NSGA-II. However, they did not present an exact method for resolution, a multi-objective mathematical model for formulation. 

Although some mathematical models have been introduced in a few studies for BHFS problem, none of them were able to capture the full granularity of the idle and blocking times. In addition, to the best of the authors' knowledge, there has been no attempt to solve BHFS with exact methods. Finally, there is a need to develop new scalable solution methods for large instances of BHFS problem due to the additional computational complexity incurred with extended details. This paper addresses these significant gaps in the literature as detailed in the following sections, starting with the MILP model formulation and formal description of the problem.







\section{Mixed Integer Linear Programming Model}\label{MILP}
In scheduling literature, several formulations are proposed for modeling the hybrid flow shop problem. In this section, we introduce a novel mixed integer linear programming model inspired by the work of \cite{initialPaper} and \cite{naderi2023mixed} which studied the HFS with the makespan criterion.

The parameters, variables, and the formulation of the mathematical model for the BHFS problem are presented in the following:\\

\begin{itemize}
\item \textbf{Sets, indices and input parameters:}
\end{itemize}
$n$ : Number of jobs,\\
$i,j$ : job index,\\
$K$ : Number of stages,\\
$k$ : Stage's index,\\
$M_k$ : Number of machines on stage $k$,\\
$m$: Machine's index,\\
$EP_k$ : Processing energy consumption at stage $k$,\\
$EI_k$ : Idle energy consumption at stage $k$,\\
$EB_k$ : Blocking energy consumption at stage $k$,\\
$P_{i,k}$ : Processing time of job $i$ at stage $k$,\\
$M$: Large integer,\\
$TPT_k$: Sum of processing time at stage $k$,\\

\begin{itemize}
\item \textbf{Decision variables:}\\
\end{itemize}
\begin{tabular}{ll}
$C_{max}$   & Total completion time \\
$TEC$            & Total energy consumption \\
$TBT$            & Total blocking time \\
$T_{idle}$       & Total idle time \\
$BT_{j,k}$       & Blocking time of job $j$ at stage $k$ \\
$BT_{j,k,m}$     & Blocking time of job $j$ on machine $m$ at stage $k$ \\
$IT_{k,m}$       & Idle time of machine $m$ at stage $k$ \\
$S_{j,k}$        & Starting time of job $j$ at stage $k$ \\
$C_{j,k}$        & Completion time of job $j$ at stage $k$ \\
$LC_{k,m}$       & Latest completion time of machine $m$ in stage $k$ \\
$ES_{k,m}$       & Earliest starting time of machine $m$ in stage $k$ \\
\end{tabular}
\begin{equation*}
    X_{i,k,m}=
    \begin{cases}
      1 & \text{if job $i$ is assigned to machine $m$ at stage $k$}, \\
      0 & \text{otherwise}.
    \end{cases}
\end{equation*}

\begin{equation*}
     Z_{i,j,k}=
    \begin{cases}
      1 & \text{if job $i$ precedes job $j$ at stage $k$}, \\
      0 & \text{otherwise}.
    \end{cases}
\end{equation*}

\begin{equation*}
    Q_{i,k,m}=
    \begin{cases}
      1 & \text{if job $i$ is the first job processed on machine $m$ at stage $k$}, \\
      0 & \text{otherwise}.
    \end{cases}
\end{equation*}

\begin{itemize}
\item \textbf{Mathematical model:}\\
\end{itemize}

\begin{equation*}
Min \ TEC 
\end{equation*}
\begin{equation*}
Min \ Cmax 
\end{equation*}

\begin{equation}\label{eq_1}
\sum_{m=1}^{M_k} X_{i,k,m} =1,\   \forall i \in \{1..n\},  \forall k \in \{1..K\}
\end{equation}

\begin{equation}\label{eq_2}
Z_{i,j,k} + Z_{j,i,k} \leq  1,\  \forall i,j \in  \{1..n\}, i\neq j , \forall k \in  \{1..K\}
\end{equation}

\begin{equation}\label{eq_3}
Z_{i,j,k} + Z_{j,i,k} \geq  X_{i,k,m} + X_{j,k,m} - 1,\  \forall i,j \in  \{1..n\}, i\neq j , \forall k \in  \{1..K\}
\end{equation}

\begin{equation}\label{eq_4}
S_{j,k} - C_{i,k} + M( 3 - X_{i,k,m} - X_{j,k,m} -Z_{i,j,k}) \geq  0,\  \forall i,j \in  \{1..n\}, i\neq j , \forall k \in  \{1..K\}
\end{equation}


\begin{equation}\label{eq_6}
C_{i,k} = S_{i,k} + P_{i,k}+BT_{i,k}\ , \forall i \in \{1..n\} , \forall k \in \{1..k\} 
\end{equation}

\begin{equation}\label{eq_7}
BT_{i,K} = 0,\ \forall i \in \{1..n\}
\end{equation}

\begin{equation}\label{eq_8}
C_{i,k} = S_{i,k+1},\  \forall i \in \{1..n\} , \forall k \in \{1..K-1\} 
\end{equation}

\begin{equation}\label{eq_9}
C_{max} \geq C_{iK}, \forall i \in \{1..n\}
\end{equation}

\begin{equation}\label{eq_10}
    BT_{ik}=S_{i,k} - C_{i,k-1} \ \forall i \in 1..n , \ \forall k \in 2..K 
\end{equation}

\begin{equation}\label{eq_11}
TBT =\sum_{i=1}^{N}\sum_{k=1}^{K-1} BT_{ik}
\end{equation}

\begin{equation}\label{eq_12}
C_{i,k} \leq LC_{k,m}  + M ( 1 - X_{i,k,m}),\  \forall i \in  \{1..n\},\forall m \in  \{1..M_{k}\}, \forall k \in  \{1..K\}
\end{equation}

\begin{equation}\label{eq_13}
S_{i,k} \geq  ES_{k,m} - M ( 1- Q_{i,k,m}),\  \forall i \in  \{1..n\},\forall m \in  \{1..M_{k}\}, \forall k \in  \{1..K\}
\end{equation}

\begin{equation}\label{eq_14}
S_{i,k} \leq  ES_{k,m} + M ( 1- Q_{i,k,m}),\  \forall i \in  \{1..n\},\forall m \in  \{1..M_{k}\}, \forall k \in  \{1..K\}
\end{equation}

\begin{equation}\label{eq_15}
    S_{i,k} \geq  ES_{k,m}- M ( 1- X_{i,k,m}), \  \forall i \in  \{1..n\},\forall m \in  \{1..M_{k}\}, \forall k \in  \{1..K\}
\end{equation}

\begin{equation}\label{eq_16}
 Q_{i,k,m} \leq   X_{i,k,m},\  \forall i \in  \{1..n\}, \forall m \in  \{1..M_k\}, \forall k \in  \{1..K\}
\end{equation}

\begin{equation}\label{eq_17}
\sum_{i=1}^{N} Q_{i,k,m}  \leq 1,\    \forall m \in  \{1..M_k\}, \forall k \in  \{1..K\}
\end{equation}

\begin{equation}\label{eq_18}
\sum_{i=1}^{N} X_{i,k,m} \leq  M \sum_{i=1}^{N} Q_{i,k,m}  ,\   \forall m \in  \{1..M_k\}, \forall k \in  \{1..K\}
\end{equation}

\begin{equation}\label{eq_19}
BT_{i,k,m} \geq  BT_{i,k} - M ( 1- X_{i,k,m}),\  \forall i \in  \{1..n\},\forall m \in  \{1..M_{k}\}, \forall k \in  \{1..K\}
\end{equation}

\begin{equation}\label{eq_20}
BT_{i,k,m} \leq  BT_{i,k} + M ( 1- X_{i,k,m}),\  \forall i \in  \{1..n\},\forall m \in  \{1..M_{k}\}, \forall k \in  \{1..K\}
\end{equation}

\begin{equation}\label{eq_21}
BT_{i,k,m} \leq   M X_{i,k,m},\  \forall i \in  \{1..n\},\forall m \in  \{1..M_{k}\}, \forall k \in  \{1..K\}
\end{equation}

\begin{equation}\label{eq_22}
\begin{split}
Idle_{k,m} =  LC_{k,m} - ES_{k,m} - \sum_{i=1}^{N} (P_{i,k}*X_{i,k,m}) - \sum_{i=1}^{N} (BT_{i,k,m}) ,\\ \forall i \in  \{1..n\},\forall m \in  \{1..M_{k}\}, \forall k \in  \{1..K\}
\end{split}
\end{equation}

\begin{equation}\label{eq_23}
Tidle = \sum_{k=1}^{K}\sum_{m=1}^{M_k} (Idle_{k,m}),\ \forall m \in  \{1..M_{k}\}, \forall k \in  \{1..K\}
\end{equation}

\begin{equation}\label{eq_24}
TEC = \sum_{i=1}^{n}\sum_{k=1}^{k} (BT_{i,k} * EB_k) + \sum_{k=1}^{K}\sum_{m=1}^{M_k} (Idle_{k,m} * EI_{k}) + \sum_{k=1}^{K} (TPT_{k} * EP_{k}) 
\end{equation}


\begin{equation} \label{eq_25}
Z_{i,j,k},Q_{i,k,m} \in \{0,1\},\  \forall i,j \in \{1..n\}, \forall k \in  \{1..K\} 
\end{equation}

\begin{equation} \label{eq_26}
X_{i,k,m},Q_{i,k,m} \in \{0,1\},\  \forall i \in \{1..n\}, \forall k \in  \{1..K\}, \forall m \in  \{1..M_{k}\}. 
\end{equation}

\subsection{Model Description}
The mathematical model aims to find the Pareto 
 optimal front that minimizes the total energy consumption $TEC$ and the latest completion time $C_{max}$.
Constraints~(\ref{eq_1}) ensure that each job is assigned to only one machine at each stage.
Constraints~(\ref{eq_2}) imply that in any stage, for any two jobs, only one can precede the other. 
Constraints ~(\ref{eq_3}) ensure that if any two jobs are assigned to the same machine at any stage, one has to precede the other.
Constraints~(\ref{eq_4}) are used to prevent overlapping of any two jobs that are assigned to the same machine on the same stage and guarantee that starting time of the successor job is later than the completion of the preceding job. 
Constraints~(\ref{eq_6}) ensure that the completion time of a job is equal to the starting time plus the processing and blocking time.
Constraints~(\ref{eq_7}) represent that there is no blocking time for any job at the last stage. 
Constraints~(\ref{eq_8}) force that the starting time 
of a given job is equal to its completion time in the previous stage.
Constraints set~(\ref{eq_9}) provides the lower bound for the makespan. 
Constraints set~(\ref{eq_10}) formulates the blocking time of a job in each stage as the difference between its completion time on that stage and the starting time on the next one.
Constraints set~(\ref{eq_11}) computes the total blocking time over all stages.
Constraints~(\ref{eq_12}) guarantee that the shutdown of each machine is later than the completion time of its last assigned job. Constraints set~(\ref{eq_13}) and (\ref{eq_14}) is the linearization of the below logical proposition and computes the machine turn-on time. 
\begin{equation*}
Q_{i,k,m} =1 \Leftrightarrow S_{i,k} =  ES_{k,m}, \  \forall i \in  \{1..n\},\forall m \in  \{1..M_{k}\}, \forall k \in  \{1..K\}
\end{equation*}

Constraints (\ref{eq_15}) provide lower bounds for $S_{i,k}$ variables and make sure the starting time is later than the turn-on time of the assigned machine.
Constraints~(\ref{eq_16}) provide an upper bound for $Q$ variables and make sure they take value only if the corresponding job is assigned to that machine. Constraints \ref{eq_17}) ensure that for each machine, only one job could be the first one processed if that machine is ever used. And constraints \ref{eq_18} guarantee that if a machine is used by any job, one of them has to be the first one so that machine turn-on time can be computed via constraints (\ref{eq_13}) and (\ref{eq_14}) as described above.


Constraints~(\ref{eq_19}) and (\ref{eq_20}) are the linearization of the following logical proposition and make sure that the blocking time of each job $i$ in each stage $k$ is equal to its blocking time on the machine $m$ that $i$ is assigned to in that stage.
\begin{equation*}
X_{i,k,m} =1 \Leftrightarrow BT_{i,k} =  BT_{i,k,m}, \  \forall i \in  \{1..n\},\forall m \in  \{1..M_{k}\}, \forall k \in  \{1..K\}
\end{equation*}
Constraints (\ref{eq_21}) give an upper bound for blocking time variables and make sure they are not getting any value on a machine $m$ in any stage $k$ if the job $i$ is not assigned to.


Constraint~(\ref{eq_22}) computes the idle time of each machine in all stages
as the time between the completion time of the last job and the starting time of the first job minus the sum of the processing and blocking times. And constraints~(\ref{eq_23}) simply sum all machine idle times and compute the total idle time.

Constraints~(\ref{eq_24}) computes the total energy consumption, $TEC$, as the sum of processing, blocking, and idle time energy consumption and constraints~(\ref{eq_25}) and ~(\ref{eq_26}) define the 
domain of the decision variables.

Size complexity of the formulated mathematical model is dominated by $\mathcal{O}(n^2 \times K)$ both for the number of variables and constraints due to variables $Z$ and constraints~(\ref{eq_2}) and ~(\ref{eq_3}).

\subsection{Model Contribution and Resolution Approach}
Several models have been introduced in the literature to address a variety of energy-efficient scheduling problems such as \cite{qin2023energy}.
However, our paper presents a novel MILP (Mixed-Integer Linear Programming) model that emerges due to its precise measurement of idle time. 
In contrast to existing models, our approach uniquely incorporates the turn-on time of each machine at the moment a job is assigned and begins processing. Furthermore, the turn-off time of the machine is equal to the completion of the last assigned job. 
These characteristics make our model unique, allowing for a more accurate representation of the production system and energy consumption and contributing to improving energy-efficient scheduling strategies while ensuring minimizing latest completion time.

There are various solution approaches for formulating objective functions of multi-objective mathematical models; lexicographic, weighted sum, goal programming, and the $\epsilon$-constraint method. In this study, we employed the augmented $\epsilon$-constraint method to exclusively derive Pareto-optimal solutions \cite{mavrotas2009effective}.
Similar to the standard $\epsilon$-constraint approach, the augmented $\epsilon$-constraint method optimizes only one of the objective functions, treating the others as constraints. However, in contrast to the classical $\epsilon$-constraint method, the augmented one defines the objective function constraints as equalities. This is accomplished by introducing necessary slack/surplus variables. Subsequently, these slack/surplus variables are incorporated into the objective function as a second term to ensure Pareto optimality.

To better understand the resolution structure, we adapted an exact approach using a constraint-based reformulation of the bi-objective model. In this formulation, one objective (e.g., total energy consumption, TEC) is optimized while the other (e.g., makespan, $C_{\text{max}}$) is transformed into a constraint with a predefined lower bound and upper bound. The two limits $(L^-, L^+)$ are obtained by dividing the two bounds by the number of intervals, which is 20 in our study.
This results in a constrained single-objective optimization model solvable using standard MILP solvers.
For instance, to minimize TEC under a makespan constraint, the modified model is formulated as follows:

\begin{align}
\text{Minimize} \quad & \text{TEC} \\
\text{Subject to:} \quad 
& \text{All standard problem constraints,} \nonumber \\
& {L}^{\text{-}} \leq C_{\text{max}} \leq {\text{L}}^{\text{+}} \label{eq:cmax_constraint}
\end{align}
By varying the value of ${\text{L}}^{\text{-}}$ and ${\text{L}}^{\text{+}}$, we can generate different trade-off solutions, effectively tracing out a portion of the Pareto front.

\subsection{An Illustrative Example}
To better explain the studied problem, we provide an illustrative simple example of a BHFS problem with two stages and two machines at each. The processing times of the five jobs as well as the energy consumption of different states are given in Table
\ref{Table_example_BHFS} in hours.

\begin{table}[t]
\caption{Processing times for the example in hours}\label{Table_example_BHFS}
\centering
\begin{tabular}{cccc}
\hline
Consumption & Job & Stage 1 & Stage 2  \\ 
\hline
Processing & 0	&	2	&	5	\\
 & 1	&	3	&	1	\\
& 2	&	7	&	5	\\
 & 3	&	2	&	9	\\
 & 4	&	4	&	2	\\
 & 5	&	5	&	1	\\
 & 6	&	7	&	7	\\
\hline 
Consumption & Processing	&	4	&	2	\\
 & Idle	&	3	&	1	\\
 & Blocking	&	2	&	3	\\
\hline
\end{tabular}
\end{table}

\begin{figure}[t]
  \centering
        \includegraphics[width=0.8\textwidth]{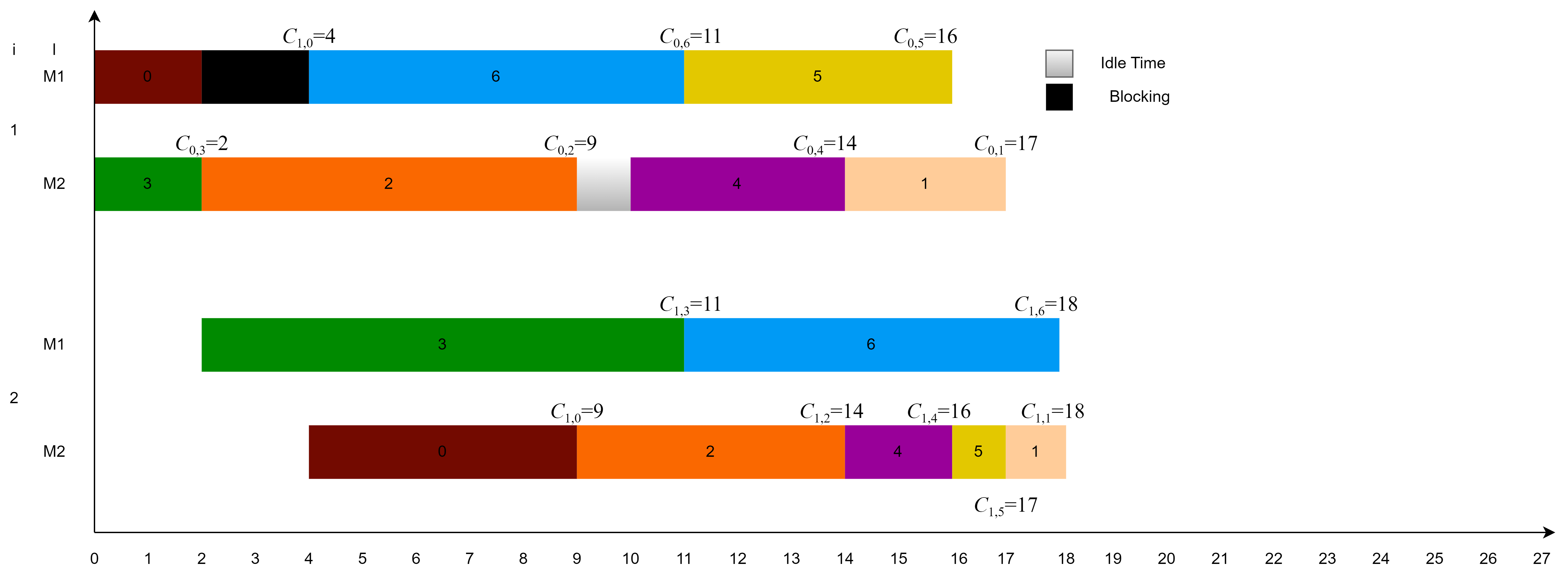}\\
  \caption{Gantt chart with minimum makespan (Cmax) }
  \label{Gant_Cmax}
\end{figure}

\begin{figure}[hbtp]
  \centering
        \includegraphics[width=0.8\textwidth]{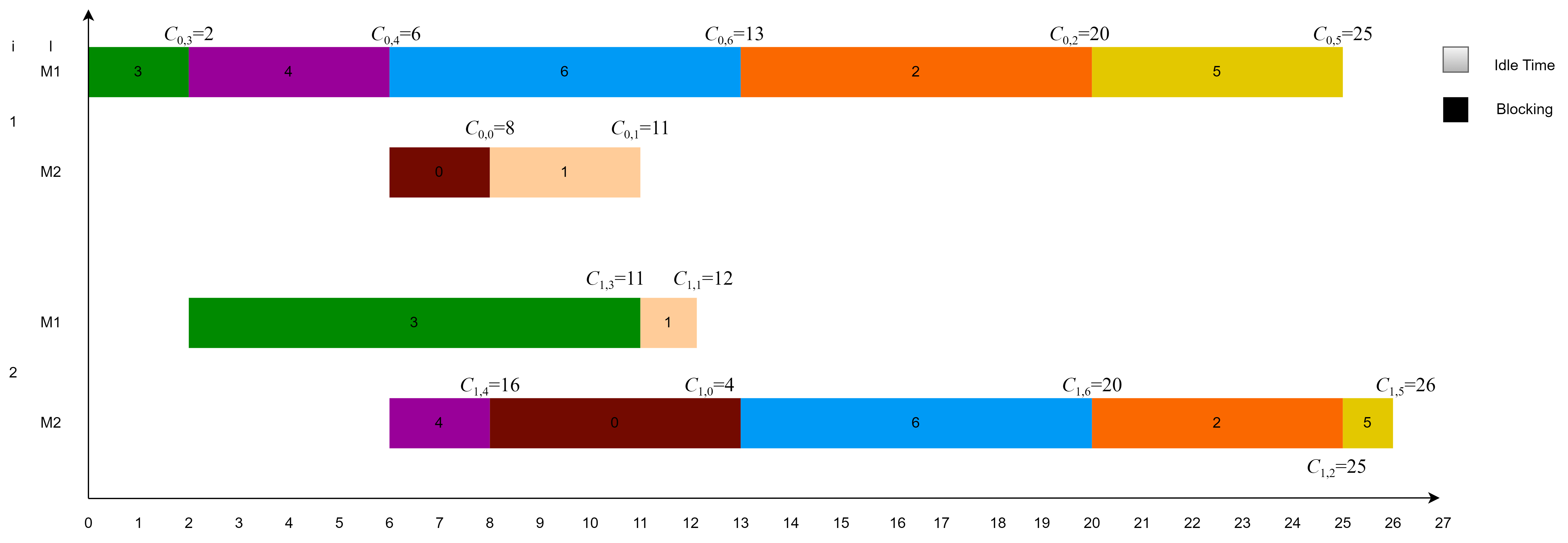}\\
  \caption{Gantt chart with minimum Total Energy Consumption (TEC)}
  \label{Gant_TEC}
\end{figure}
Using the MILP model, we solve the aforementioned example by considering the two distinct objective functions. The Gantt chart depicted in Figure \ref{Gant_Cmax} illustrates the results for the Makespan criterion, giving Cmax of 18 and TEC of 191. While, Figure \ref{Gant_TEC} presents the exact resolution for the TEC criterion, resulting in a TEC value of 172 and a corresponding Cmax of 26.

\section{Refined Iterated Pareto Greedy}\label{R-MOIG}
\begin{figure}[t]
  \centering
        \includegraphics[width=0.6\textwidth]{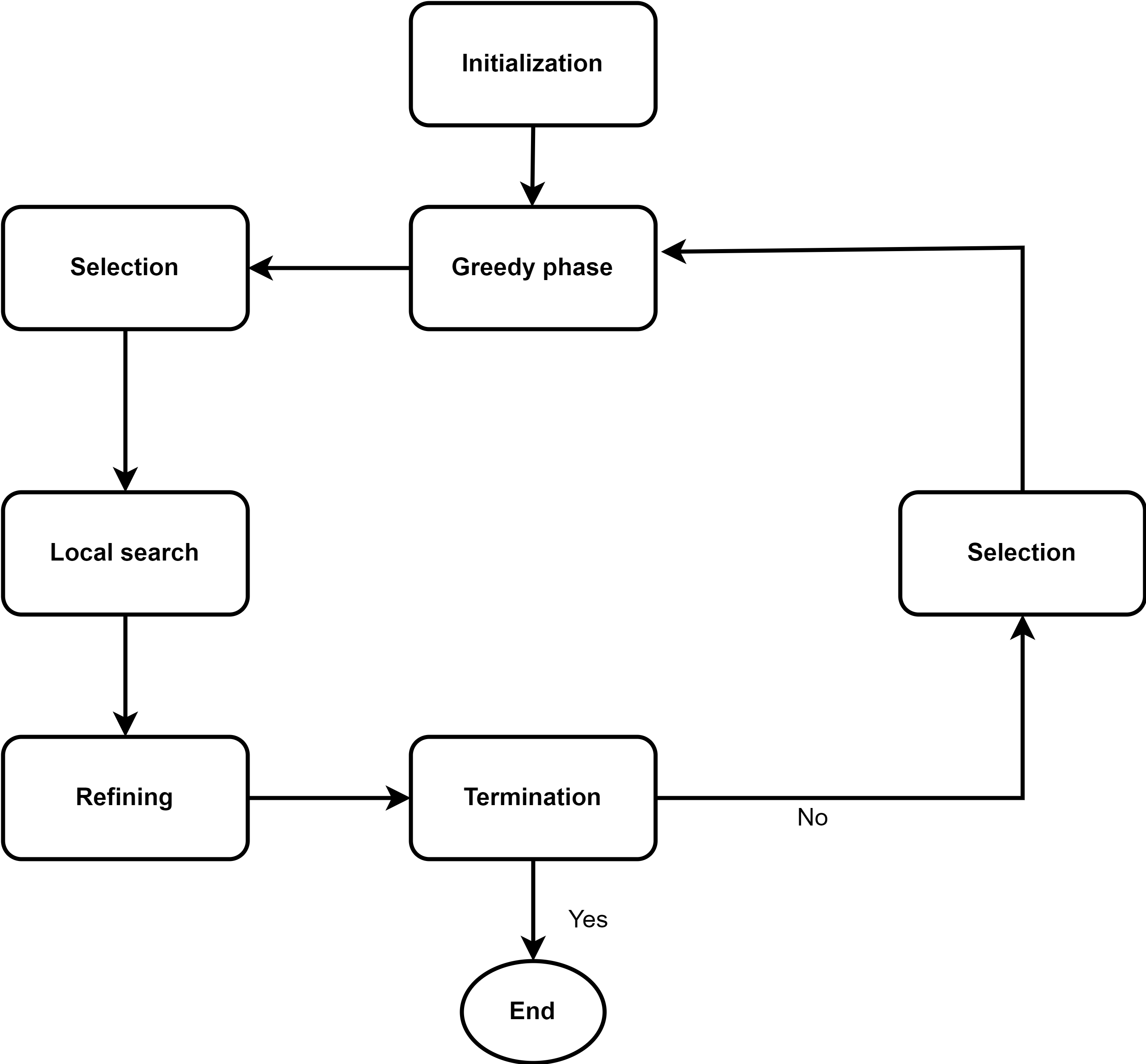}\\
  \caption{Flowchart of the proposed R-IPG}
  \label{Flow_chart}
\end{figure}

In the rich multi-objective optimization literature, several algorithms have been introduced to solve scheduling problems.
In this work, we developed an improved multi-objective iterated greedy (MOIG) to solve the energy-efficient BHFS.
Iterated greedy (IG) was introduced for the first time by \cite{ruiz2007simple} and has appeared a high performance in solving scheduling problems. 
In its basic form, IG consists of a set of successive partial destruction and reconstruction of a solution in order to increase diversification, the obtained result could be improved by a local search operator.
The first adaptation of IG to solve multi-objective problem starts with the work of \cite{framinan2008multi}. In this paper, the authors proposed an extended IG based on initialization and greedy phase to solve the flowshop scheduling problem aiming to get the Pareto front of Makespan and total flow time.
In \cite{minella2011restarted} a new variant of MOIG iterated was developed to solve the multi-objective flowshop with makespan, tardiness, and flowtime criteria.
Later, many versions of MOIG were introduced in the literature, and all of them produced very high results quality which represents a motivation to choose this method in our work.
The proposed algorithm is referred to as refined Iterated Pareto Greedy or (R-IPG). The remainder of this section is dedicated to describing in detail the different operators of our proposal. 
Briefly, the R-IPG includes five operators (Initialization, selection, greedy phase, local search, and Refining phase). Moving from one operator to another requires a selection phase capable of identifying the most suitable candidate at each iteration, instead of processing all solutions in the Pareto front. The main objective is to identify the most isolated solution and attempt to replace it with a new one.
Also, the novel refining phase, one of the contributions of this study, can improve the quality of the Pareto front by addressing each solution individually.
Figure \ref{Flow_chart} shows all steps of the proposed R-IPG which are detailed in the following subsections.

\subsection{Initialization}
As with the single objective IG, the quality of the initial solution in a multi-objective algorithm has an impact on its performance. In this work, we are dealing with two objectives, therefore, one solution is initialized for each one.
The NEH heuristic of \cite{nawaz1983heuristic} is used 
to generate a high-quality solution for the Makespan criterion. This dispatching rule is among the best for this criterion and it has been used in several studies when dealing with Makespan.

For the second objective, we used the modified NEH heuristic as a dispatching rule described in \cite{missaoui2023energy}, in which the only modification is changing the criterion.
Jobs are sorted according to their sum of processing time. Starting with the initial partial sequence including only the first job and then inserting each job in all possible positions and keeping the insertion that provides the minimum TEC. The process is repeated until we insert all jobs.
In conclusion, two solutions are obtained to constitute the initial Pareto front before proceeding to the next step, there is no guarantee that these two solutions will be in the final Pareto front.

\subsection{Selection Phase}

Moving from one operator to another, a set of non-dominated solutions is achieved. The idea of processing all of them doesn't look promising in terms of time and diversity. Therefore, only one solution will be selected to be handled with the next operator.
A completely random selection appeared inadequate to cover all regions of the Pareto front. For this reason, 
we introduced a hybrid selection. The idea is to calculate the crowding distance (CD) for each solution as introduced in \cite{deb2002fast} and the solution with the higher CD will be selected. This could help to choose the most isolated solution in order to reduce the gap in the resulting Pareto front. This could be useful if there is an improvement in the Pareto front but otherwise, it results in immature
 convergence and low global performance. To skip shapes, a random sequence will be selected if the current Pareto has not changed. 
\subsection{Greedy Phase}
This operator is used to destroy and reconstruct one selected solution from the Pareto front aiming to generate a set of solutions.
The first stride is to remove $d$ random jobs from the initial solution $\pi$ and put them in a new set called $D$. We mention that $d$ is a parameter that must be calibrated.
Once the destruction is finished, The construction step starts when each job from $D$ is reinserted in the partial sequence.
The process differs from the original IG in that it involves reinserting each job in all possible positions and then selecting the position that yields the best fitness for retention.
In the R-IPG construction phase, we insert the first job of $D$ in all possible positions, at each insertion the new partial sequence is saved in a set $NWS$. The rest of the jobs of $D$ are reinserted in the same way in all solutions of the set $NWS$. To avoid the growing number of generated solutions, The dominated partial sequences are removed from $NWS$ after each insertion and only the non-dominated sequences are kept for the next insertions. The main steps of this greedy phase are given in the Algorithm. 
\ref{alg:greedyphase}

\begin{algorithm}[t]
\caption{Greedy Phase}
\label{alg:greedyphase}
\begin{algorithmic}[1]
\Require $\pi$, $d$
\Ensure $NDS$

\State Remove $d$ jobs at random from $\pi$ and put them in set $D$
\State $NDS \gets \{\pi\}$

\For{$i = 1$ to $d$}
    \State $NWS \gets \emptyset$
    \For{Partial sequence $\pi$ in $NDS$}
        \For{each position $j$ in $\pi$}
            \State $Seq_{(\pi,i,j)} \gets$ insertion of job $i$ in position $j$
            \State add $Seq_{(\pi,i,j)}$ to $NWS$
        \EndFor
        \State $NDS \gets$ non-dominated solutions from $NWS$
    \EndFor
\EndFor
\end{algorithmic}
\end{algorithm}

\subsection{Local Search}
After the greedy phase, one solution from the current Pareto front is selected to move to the local search.
The main outline of this approach is given in Algorithm \ref{alg:localsearch}. 
Briefly, the principle of this operator is simple, one job is removed from the sequence and reinserted in all possible positions. At each insertion, the newly generated sequence is saved in a set $NPS$ and the number of solutions will be equal to the number of jobs.

\begin{algorithm}[t]
\caption{Local Search}
\label{alg:localsearch}
\begin{algorithmic}[1]
\Require $\pi$
\Ensure $NDS$

\State Remove one job at random from $\pi$
\For{each position $j$ in $\pi$}
    \State $Seq_{(\pi,i,j)} \gets$ insertion of $i$ in position $j$
    \State add $Seq_{(\pi,i,j)}$ to $NWS$
\EndFor

\State $NDS \gets$ non-dominated solutions from $NWS$
\end{algorithmic}
\end{algorithm}

\subsection{Refining}
After the local search phase, we move to the last step of our approach which aims to improve the quality of the current Pareto front.
The approach involves the application of perturbations on each solution individually.
A replacement of the current solution is performed only when the newly generated solution dominates the old one. The Refining operator is based on insertion and interchange. These two distinct neighborhood structures are sequentially applied $Loop\_Size$ times to each solution in the front. The main outline of the refining technique is given in Algorithm. \ref{alg:Refining}. In some cases, the output could be strictly a new set of solutions including dominated and no-dominated. To deal with this, a non-dominated sorting is applied aiming to remove dominated solutions.
The output of this tool represents the final Pareto front of the iteration that will move to the greedy phase. We must mention that $Loop\_Size$ is a parameter that must be calibrated.

\begin{algorithm}[t]
\caption{Refining Phase}
\label{alg:Refining}
\begin{algorithmic}[1]
\Require $P$
\Ensure $P'$

\For{each $\pi$ in $P$}
    \State loop $\gets 0$
    \While{loop $<$ loop\_max}
        \State $N\_counter \gets 1$
        \While{$N\_counter < 3$}
            \If{$N\_counter = 1$}
                \State $\pi' \gets \text{Insertion movement}(\pi)$
            \ElsIf{$N\_counter = 2$}
                \State $\pi' \gets \text{Interchange movement}(\pi)$
            \EndIf
            \If{$\pi'$ dominates $\pi$}
                \State $\pi \gets \pi'$
                \State $N\_counter \gets 1$
            \Else
                \State $N\_counter \gets N\_counter + 1$
            \EndIf
        \EndWhile
        \State loop $\gets$ loop + 1
    \EndWhile
\EndFor

\State $P' \gets$ non-dominated solutions from $P$
\end{algorithmic}
\end{algorithm}

\section{Computational Evaluation and Statistical Experimentation} \label{CompEXP}
This section is dedicated to evaluating the performance of the proposed R-IPG method. In the first way, the generated benchmark as well as the performance indicators are described. Later, a general idea is given about the two competing methods. Finally, we provide the  Computational comparisons. 

\subsection{Benchmark, Metrics and  Experimental Setting}
To address the effectiveness of the proposed method against competing methods, we need to generate a benchmark of instances covering all possible characteristics of an instance. 
Therefore, three parameters are utilized: the number of jobs ($n$), the number of stages ($g$), and the number of machines per stage ($m_{i}$). 
Based on $n$, we distinguish three different sizes, the small instances when ($n$ = 6,8,10,12), Medium instances ($n = 15,20,30$), and large instances ($n$= 50,100).
In total, 9 classes of $n$, each one which is generated 6 times adopting the combination of ($g = 2,3,4$) and ($m_{i} = 2,3$). The processing time is randomly generated from a uniform distribution [1..99] while the energy consumption values at each stage are generated from uniform distributions of [1..3], [5..7], and [3..5] for processing energy, blocking energy and idle energy respectively. In conclusion, our benchmark encompasses a total of 54 instances eligible to test the performance of our proposal. 
For the calibration benchmark, we generated a set equal to 50\% of test instances. Each instance is formed by randomly selecting one value of $M_k$ from each set of $n$ and $K$ values previously utilized.

Unlike single objective cases, the comparison between outcomes of the different algorithms is not trivial. However, each one of them provides a set of non-dominated solutions and the comparison needs specific metrics. In this paper, we adopt two performance indicators.
The first is the hypervolume indicator ($I_h$) introduced in \cite{zitzler1999multiobjective} and widely used in the literature. In bi-objective optimization, $I_h$ is used to measure the area covered by the non-dominated solution of the Pareto front. All objective values are normalized before calculating the $I_h$.  The reference point needed to close the space is set to $(1.2, 1.2)$. Therefore, the maximum value could be 1.44. The higher value of $I_h$ reflects a higher performance of a given method.\\

Formally, the $I_h$ metric can be defined as follows: let $\mathcal{P}$ be the set of Pareto fronts and $P$ $\in$ $\mathcal{P}$ is the set of solutions generated by a given method.
$N_{Obj}$ is the number of objective criteria and $N_{Sol}$ is the number of solutions in $P$ where $pt_{i,j}$ indicates each objective value for criteria $j$ in solution $i$. Then the hypervolume for $P$ can be calculated as:\\
$I_h(P)=\sum\limits_{i=1}^{N_{Sol}} \sum\limits_{j=1}^{N_{Obj}} ((pt_{i,j} - 1.2 * min_j \mathcal{P}) \div (1.2 * (max_j \mathcal{P} - min_j \mathcal{P}))$\\ 
where $min_j \mathcal{P}$ and $max_j \mathcal{P}$ are the best and worst values, respectively, for objective criteria $j$ over all the frontiers in $\mathcal{P}.$

The second used metric is a popular performance indicator called Generational distance ($GD$) \cite{ishibuchi2015modified}.
It provides a measure of how far the solutions in a particular generation are from the global Pareto front
or reference set of solutions.
More formally, the $GD$ indicator can be defined as follows:\\
$GD(P)= (1 \div N_{Sol}) \times \sqrt{\sum\limits_{j=1}^{N_{Sol}} d_{j}^{2}}$ where $d_{j}^{2}$ is the Euclidian distance from the solutions to their nearest reference point.

The lower value of GD indicates higher effectiveness. 
The use of these two indicators provides a general idea about the quality of the Pareto front.

The instances in this study were addressed using the augmented epsilon constraint tool, where the first objective was to minimize the Makespan taking into account the TEC as a constraint. 
To achieve this, we began by determining the permissible ranges for each individual objective using lexicographic optimization aiming to get the payoff table. 
By reason of the NP-hard nature of the BHFS problem, we generated non-dominated solution sets for all instances using an evenly higher $\epsilon$ level, which was computed by segmenting the TEC objective function's range into 20 uniform grids. A time limit of three minutes was given to each iteration resulting in a total of 1 hour for every instance.

The proposed IG, the NSGA-II, and MOIG are coded with C++ in an integrated development environment (IDE) Microsoft Visual Studio 22. while the augmented epsilon constraints are coded using ILOG Concert Technology of IBM Cplex 20.0.

All experiments are conducted on a Personal Computer with Microsoft Windows 10 equipped with a processor Intel Core i7 and 16 GB RAM. 
Each method is tested independently 10
times, to guarantee fair competition, the stopping criterion for all algorithms depends on the size of the instance as $CPU = n * g * 200$ which gives 8 seconds for an instance of 10 jobs and 4 stages. 

\subsection{Competing Methods and Calibration}
In order to test the performance of the proposed IG and the introduced MILP, two approaches well selected from the literature are used as competing methods. 
The first algorithm is the Non-dominated Sorting Genetic Algorithm II (NSGA-II) 
introduced by \cite{deb2002fast}. It is a widely used evolutionary algorithm for solving multi-objective optimization problems. 
 NSGA-II  is a widely used evolutionary algorithm that operates on a population of solutions. It uses genetic operators (selection, crossover, and mutation) along with a fast non-dominated sorting and crowding distance mechanism to maintain a diverse and converging set of Pareto-optimal solutions.
The second method used is the multi-objective iterated greedy (MOIG) introduced in \cite{missaoui2023energy}.
MOIG is the first extension of the original IG introduced by \cite{framinan2008multi}
to handle multi-objective optimization problems.
MOIG is based on a destruction and construction heuristic. It iteratively removes and reconstructs parts of the current solution to explore the search space, balancing exploitation and diversification.

\begin{figure}
    \begin{minipage}{0.5\textwidth}
        \centering
        \includegraphics[width=\linewidth]{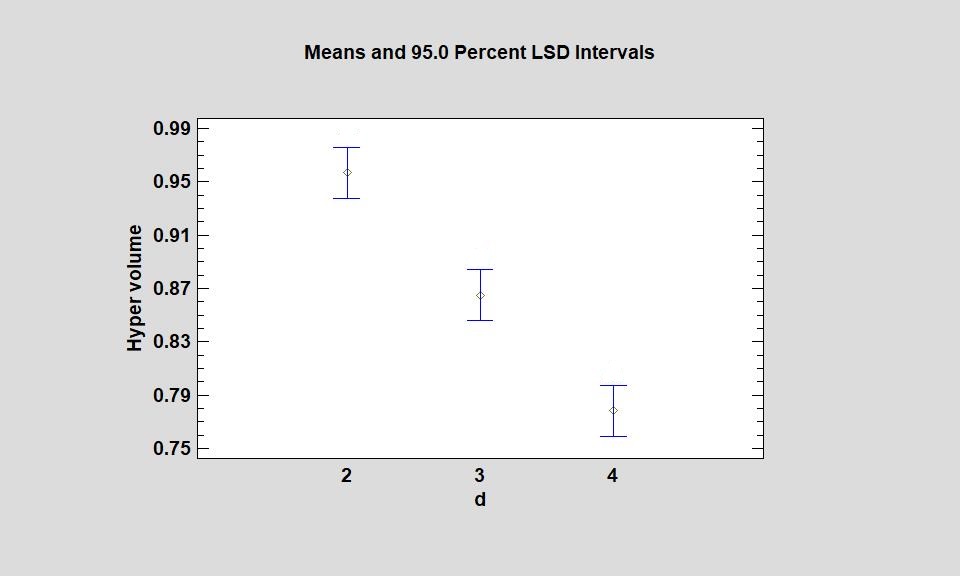}
        \label{fig:figure12}
    \end{minipage}%
    \begin{minipage}{0.5\textwidth}
        \centering
        \includegraphics[width=\linewidth]{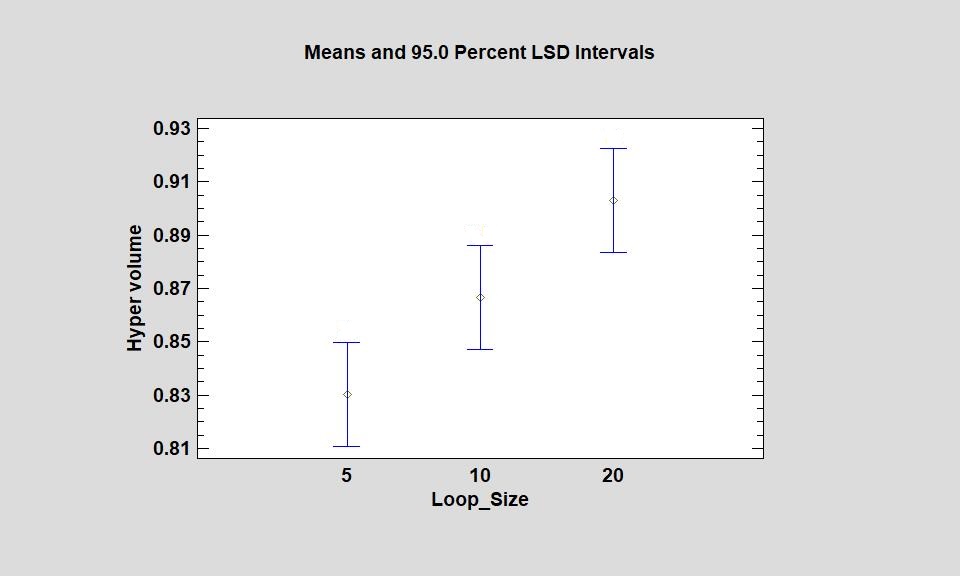}
        \label{fig:figure25}
    \end{minipage}
    \caption{{$I_h$} means plots for the R-IPG parameters}
    \label{DOE_HV}
\end{figure}

\begin{figure}
    \begin{minipage}{0.5\textwidth}
        \centering
        \includegraphics[width=\linewidth]{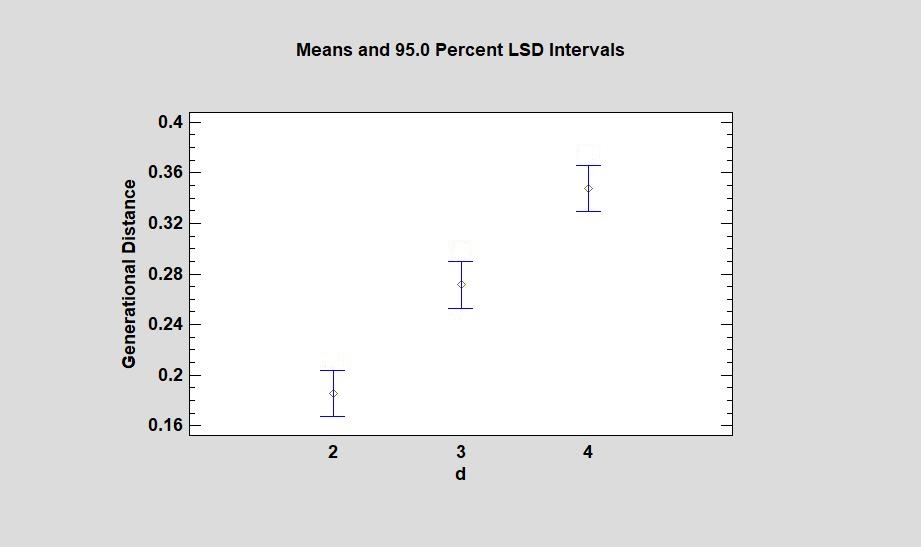}
        \label{fig:figure13}
    \end{minipage}%
    \begin{minipage}{0.5\textwidth}
        \centering
        \includegraphics[width=\linewidth]{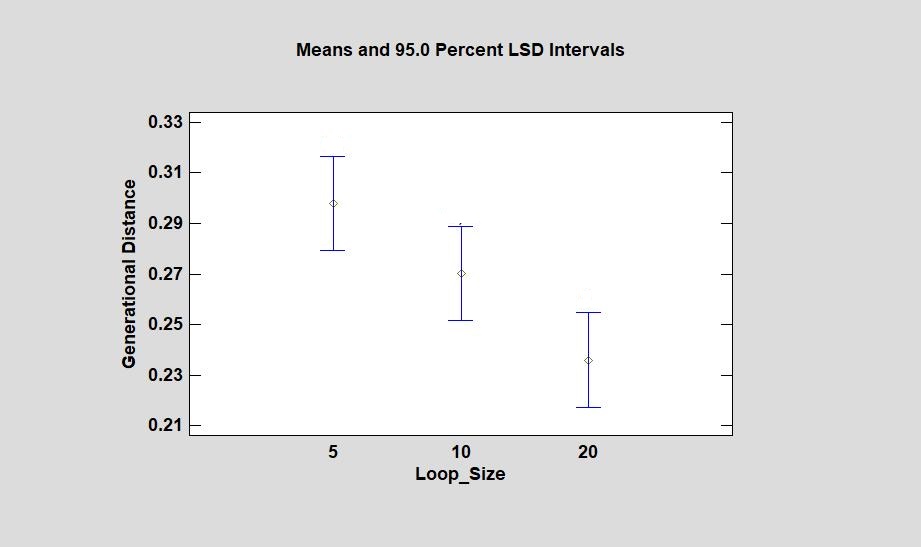}
        \label{fig:figure21}
    \end{minipage}
    \caption{{$GD$} means plots for the R-IPG parameters}
    \label{DOE_GD}
\end{figure}

The R-IPG algorithm is tailored more specifically to the problem under study and uses an intensification strategy guided by the selection operator and the refining phase
To increase the efficiency of a given method, a calibration process is essential to select the best parameter set.
Our proposed R-IPG method depends on two parameters:
\begin{itemize}
    \item Parameter (d) or the destruction size refers to the number of removed jobs during the destruction phase. This parameter is tested at three levels (2, 3, 4).
    \item Parameter (loop\_size) refers to the number of iterations during the refining phase. This parameter is also tested at three levels (5, 10, and 20).
\end{itemize}  
We propose a Design of Experiments (DOE) with all levels aiming to evaluate the different algorithm configurations.
Each configuration is repeated 5 times with the 27 calibration instances to give 3 $\times$ 3 $\times$ 5 $\times$ 27 experiments. In total, 1215 Pareto fronts are obtained. 
The same stopping criterion as experiments is used. The hypervolume indicator ($I_h$) and the  Generational distance $(GD)$ are employed as the response variables.
To statistically compare all possible configurations, we applied the analysis of variance (ANOVA).
The results show that the different analyses are statistically significant.
Figures \ref{DOE_HV} and \ref{DOE_GD} show the mean plots of the two parameters (d) and (Loop\_size) for $I_h$ and $GD$ respectively.

\subsection{Computational Results}
After realizing all experiments, we compare the introduced R-IPG approach with two competing methods and the augmented epsilon constraint (AUG) in the three different sizes of instances. Testing the proposal with small, medium, and large instances could give a general insight into its performance.

\begin{table}

\caption{$I_h$ and $GD$ values for small instances (Best values are in bold)}\label{Results_small}

\tiny
\centering
\begin{tabular}{ccccccccccc}

 &  \multicolumn{3}{c}{}& \multicolumn{2}{c}{$I_h$} & \multicolumn{5}{c}{GD} \\
\hline
n	&	g	&	m	&	NSGA	&	 MOIG	&	RIPG	&	Aug	&	NSGA	&	MOIG	&	RIPG	&	Aug	\\
\hline
6	&	2	&	2	&	0.93	&	0.96	&	0.96	&	\textbf{1.36}	&	0.357	&	0.346	&	0.333	&	\textbf{0.033}	\\
	&		&	3	&	0.76	&	0.75	&	0.74	&	\textbf{1.42}	&	0.707	&	0.735	&	0.709	&	\textbf{0.007}	\\
	&	3	&	2	&	0.34	&	0.34	&	0.34	&	\textbf{1.34}	&	0.698	&	0.697	&	0.707	&	\textbf{0.084}	\\
	&		&	3	&	0.95	&	0.94	&	0.93	&	\textbf{1.37}	&	0.398	&	0.432	&	0.329	&	\textbf{0.037}	\\
	&	4	&	2	&	0.90	&	0.92	&	0.92	&	\textbf{1.39}	&	0.662	&	0.594	&	0.546	&	\textbf{0.014}	\\
	&		&	3	&	0.85	&	0.85	&	0.87	&	\textbf{1.43}	&	0.598	&	0.614	&	0.565	&	\textbf{0.014}	\\
Av	&		&		&	0.79	&	0.79	&	0.79	&	\textbf{1.39}	&	0.570	&	0.570	&	0.532	&	\textbf{0.031}	\\
\hline
8	&	2	&	2	&	1.18	&	1.26	&	1.24	&	\textbf{1.40}	&	0.479	&	0.419	&	0.455	&	\textbf{0.009}	\\
	&		&	3	&	0.98	&	1.04	&	1.07	&	\textbf{1.42}	&	0.590	&	0.600	&	0.479	&	\textbf{0.005}	\\
	&	3	&	2	&	0.87	&	0.92	&	0.86	&	\textbf{1.35}	&	0.479	&	0.439	&	0.419	&	\textbf{0.051}	\\
	&		&	3	&	0.71	&	0.75	&	0.80	&	\textbf{1.40}	&	0.555	&	0.583	&	0.537	&	\textbf{0.054}	\\
	&	4	&	2	&	0.83	&	0.85	&	0.82	&	\textbf{1.42}	&	0.390	&	0.363	&	0.390	&	\textbf{0.051}	\\
	&		&	3	&	0.85	&	0.83	&	0.84	&	\textbf{1.44}	&	0.638	&	0.661	&	0.651	&	\textbf{0.326}	\\
Av	&		&		&	0.90	&	0.94	&	0.94	&	\textbf{1.41}	&	0.522	&	0.511	&	0.489	&	\textbf{0.082}	\\
10	&	2	&	2	&	0.77	&	0.68	&	0.85	&	\textbf{1.41}	&	0.483	&	0.531	&	0.333	&	\textbf{0.018}	\\
\hline
	&		&	3	&	0.92	&	0.91	&	1.01	&	\textbf{1.42}	&	0.371	&	0.405	&	0.306	&	\textbf{0.022}	\\
	&	3	&	2	&	0.98	&	1.04	&	1.03	&	\textbf{1.39}	&	0.175	&	0.157	&	0.136	&	\textbf{0.121}	\\
	&		&	3	&	0.85	&	0.73	&	0.85	&	\textbf{1.39}	&	0.450	&	0.480	&	0.393	&	\textbf{0.021}	\\
	&	4	&	2	&	0.86	&	1.00	&	0.90	&	\textbf{1.30}	&	0.345	&	0.327	&	0.354	&	\textbf{0.006}	\\
	&		&	3	&	0.88	&	0.49	&	0.89	&	\textbf{1.44}	&	0.583	&	0.892	&	0.520	&	\textbf{0.004}	\\
Av	&		&		&	0.88	&	0.81	&	0.92	&	\textbf{1.39}	&	0.401	&	0.465	&	0.340	&	\textbf{0.032}	\\
\hline
12	&	2	&	2	&	1.22	&	1.31	&	1.27	&	\textbf{1.33}	&	0.087	&	\textbf{0.046}	&	0.050	&	0.183	\\
	&		&	3	&	1.00	&	0.98	&	1.16	&	\textbf{1.40}	&	0.271	&	0.350	&	0.184	&	\textbf{0.059}	\\
	&	3	&	2	&	0.81	&	0.99	&	1.07	&	\textbf{1.08}	&	0.247	&	0.190	&	0.147	&	\textbf{0.058}	\\
	&		&	3	&	1.05	&	0.89	&	1.14	&	\textbf{1.44}	&	0.539	&	0.615	&	0.382	&	\textbf{0.119}	\\
	&	4	&	2	&	0.90	&	1.23	&	\textbf{1.25}	&	1.13	&	0.348	&	\textbf{0.059}	&	0.080	&	0.100	\\
	&		&	3	&	0.67	&	0.60	&	0.82	&	\textbf{1.32}	&	0.257	&	0.293	&	0.104	&	\textbf{0.004}	\\
Av	&		&		&	0.94	&	1.00	&	1.12	&	\textbf{1.28}	&	0.291	&	0.259	&	0.158	&	\textbf{0.087}	\\
\hline
G.Avg	&		&		&	0.88	&	0.89	&	0.94	&	\textbf{1.37}	&	0.446	&	0.451	&	0.380	&	\textbf{0.058}	\\
\hline 
\end{tabular}
\end{table}

Table \ref{Results_small} shows the hypervolume indicator $I_h$ and the generational distance $GD$ values for the small benchmark arranged by the number of $Jobs$ and $Stages$. As seen, the AUG 
returns the best values for both metrics with an average $I_h = 1.37$ and $GD = 0.058$ which means that most instances are optimally solved.

\begin{figure}
    \begin{minipage}{0.5\textwidth}
        \centering
        \includegraphics[width=\linewidth]{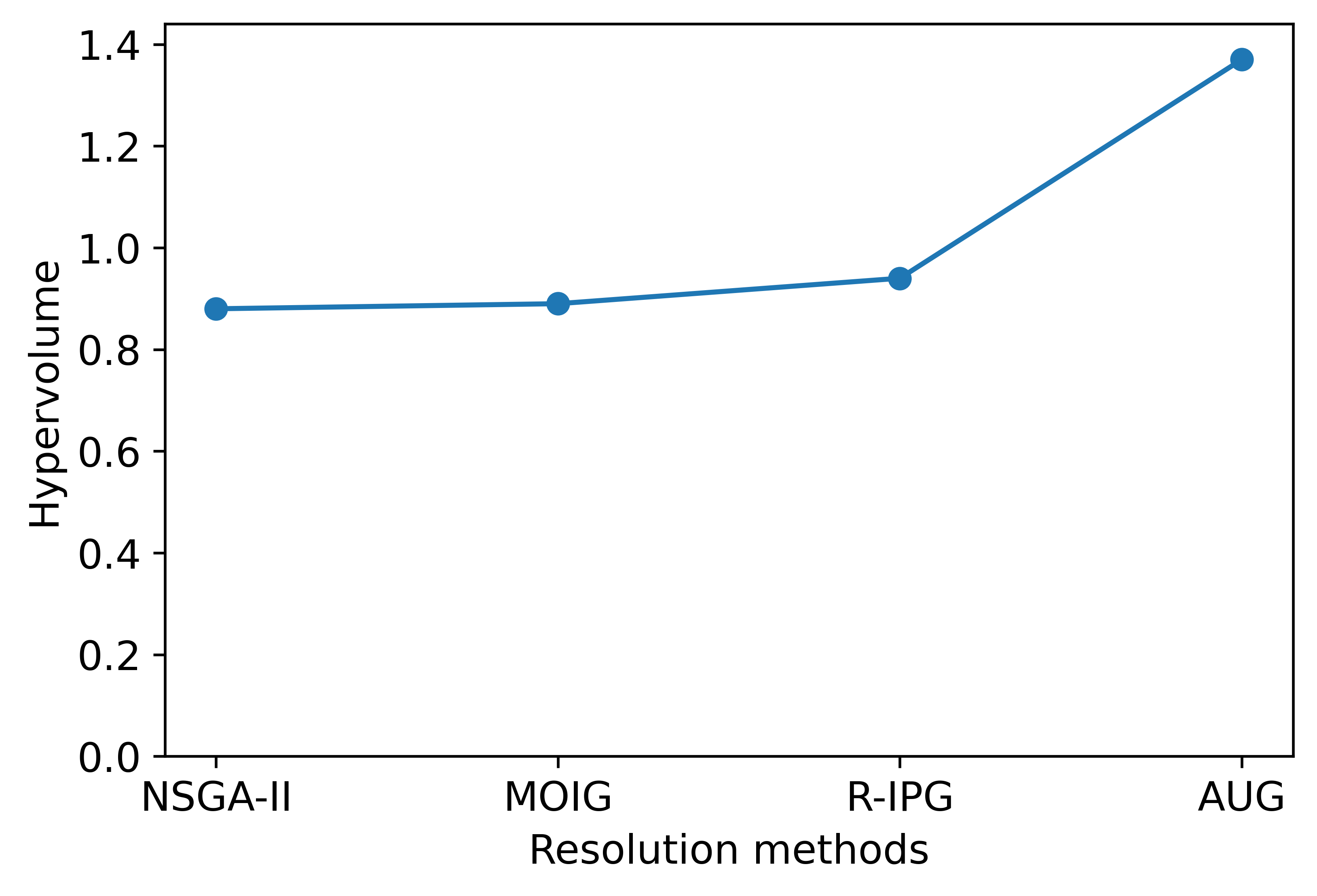}
        \label{fig:figure15}
    \end{minipage}%
    \begin{minipage}{0.5\textwidth}
        \centering
        \includegraphics[width=\linewidth]{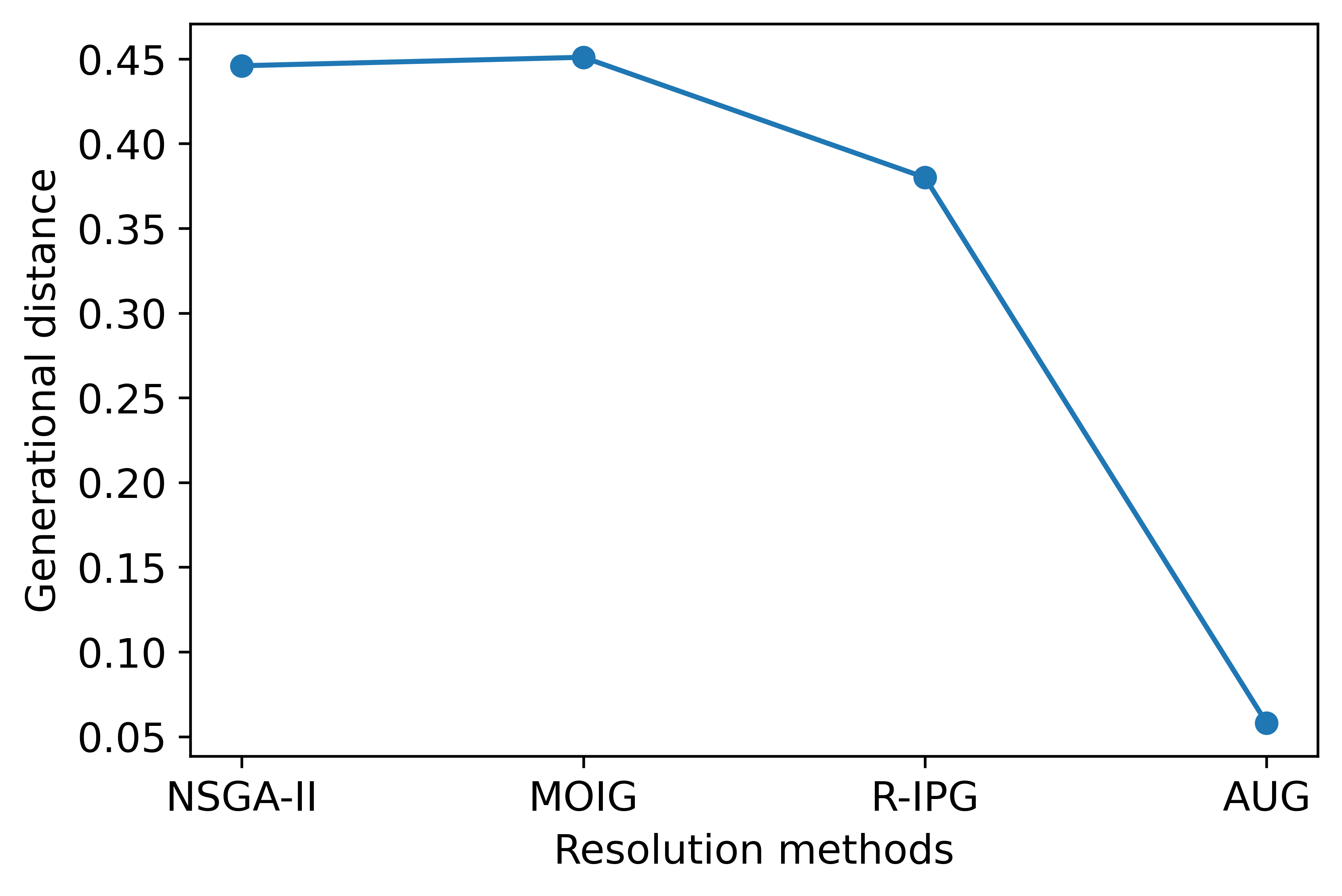}
        \label{fig:figure22}
    \end{minipage}
    \caption{$I_h$ and $GD$ means plots for all tested methods with small instances}
    \label{means_plots_S}
\end{figure}

As seen in Figure~\ref{means_plots_S}, the proposed R-IPG algorithm yields results that are closest to the best-known solutions, with an average hypervolume indicator of $I_h = 0.94$ and a Generational Distance of $GD = 0.380$. These values demonstrate the strong convergence capability of R-IPG toward the reference Pareto front, as well as its ability to maintain a high-quality distribution of solutions in the objective space, especially for small-sized problem instances.

The results provided by MOIG and NSGA-II with small instances are close in both criteria and there is no clear dominance.

The medium benchmark in our work includes instances with 15,20, and 20 jobs. The average results are given in Table \ref{Results_medium}. We can observe that the results differ significantly for small instances. The AUG exhibits its optimal performance only when $n=15$, while it deteriorates with the remainder of instances.

\begin{table}

\caption{$I_h$ and $GD$ values for medium instances (Best values are in bold)}\label{Results_medium}

\tiny
\centering
\begin{tabular}{ccccccccccc}

 &  \multicolumn{3}{c}{}& \multicolumn{2}{c}{$I_h$} & \multicolumn{5}{c}{GD} \\
\hline
n	&	g	&	m	&	NSGA	&	 MOIG	&	R-IPG	&	Aug	&	 NSGA	&	MOIG	&	 R-IPG	&	Aug	\\
\hline
15	&	2	&	2	&	0.43	&	1.03	&	1.15	&	\textbf{1.20}	&	0.700	&	0.207	&	0.072	&	\textbf{0.001}	\\
	&		&	3	&	0.73	&	0.82	&	1.12	&	\textbf{1.37}	&	0.396	&	0.322	&	\textbf{0.105}	&	0.125	\\
	&	3	&	2	&	0.80	&	1.14	&	\textbf{1.25}	&	1.04	&	0.295	&	0.157	&	\textbf{0.083}	&	0.224	\\
	&		&	3	&	0.65	&	0.79	&	1.15	&	\textbf{1.25}	&	0.602	&	0.518	&	0.211	&	\textbf{0.093}	\\
	&	4	&	2	&	0.73	&	1.13	&	1.14	&	\textbf{1.30}	&	0.492	&	0.185	&	0.120	&	\textbf{0.051}	\\
	&		&	3	&	0.83	&	0.65	&	1.15	&	\textbf{1.26}	&	0.330	&	0.483	&	0.157	&	\textbf{0.066}	\\
Av	&		&		&	0.69	&	0.93	&	1.16	&	\textbf{1.24}	&	0.469	&	0.312	&	0.125	&	\textbf{0.094}	\\
\hline
20	&	2	&	2	&	0.38	&	1.10	&	\textbf{1.22}	&	1.07	&	0.793	&	0.219	&	\textbf{0.105}	&	0.211	\\
	&		&	3	&	0.48	&	0.64	&	\textbf{1.20}	&	1.10	&	0.630	&	0.479	&	0.137	&	\textbf{0.131}	\\
	&	3	&	2	&	0.38	&	1.06	&	\textbf{1.22}	&	0.54	&	0.751	&	0.216	&	\textbf{0.078}	&	0.391	\\
	&		&	3	&	0.51	&	0.57	&	\textbf{1.27}	&	0.59	&	0.731	&	0.643	&	\textbf{0.084}	&	0.507	\\
	&	4	&	2	&	0.49	&	1.06	&	\textbf{1.19}	&	0.55	&	0.729	&	0.245	&	\textbf{0.148}	&	0.671	\\
AV	&		&		&	0.45	&	0.89	&	\textbf{1.22}	&	0.77	&	0.727	&	0.361	&	\textbf{0.111}	&	0.382	\\
\hline
30	&	2	&	2	&	0.34	&	0.89	&	\textbf{1.29}	&	0.69	&	0.769	&	0.306	&	\textbf{0.069}	&	0.462	\\
	&		&	3	&	0.54	&	0.95	&	\textbf{1.05}	&	0.57	&	0.809	&	0.433	&	\textbf{0.376}	&	0.743	\\
	&	3	&	2	&	0.52	&	0.85	&	\textbf{1.20}	&	0.33	&	0.724	&	0.386	&	\textbf{0.138}	&	1.003	\\
	&		&	3	&	0.40	&	0.71	&	\textbf{1.15}	&	0.64	&	0.763	&	0.473	&	\textbf{0.149}	&	0.703	\\
	&	4	&	2	&	1.11	&	1.27	&	\textbf{1.37}	&	0.58	&	0.184	&	0.085	&	\textbf{0.025}	&	0.916	\\
	&		&	3	&	1.07	&	1.03	&	\textbf{1.25}	&	0.17	&	0.280	&	0.285	&	\textbf{0.134}	&	1.210	\\
Av	&		&		&	0.72	&	0.96	&	\textbf{1.22}	&	0.45	&	0.544	&	0.322	&	\textbf{0.146}	&	0.893	\\
\hline
G.Avg	&		&		&	0.64	&	0.93	&	\textbf{1.20}	&	0.80	&	0.570	&	0.329	&	\textbf{0.129}	&	0.485	\\

\hline 
\end{tabular}
\end{table}

\begin{figure}[t]
    \begin{minipage}{0.5\textwidth}
        \centering
        \includegraphics[width=\linewidth]{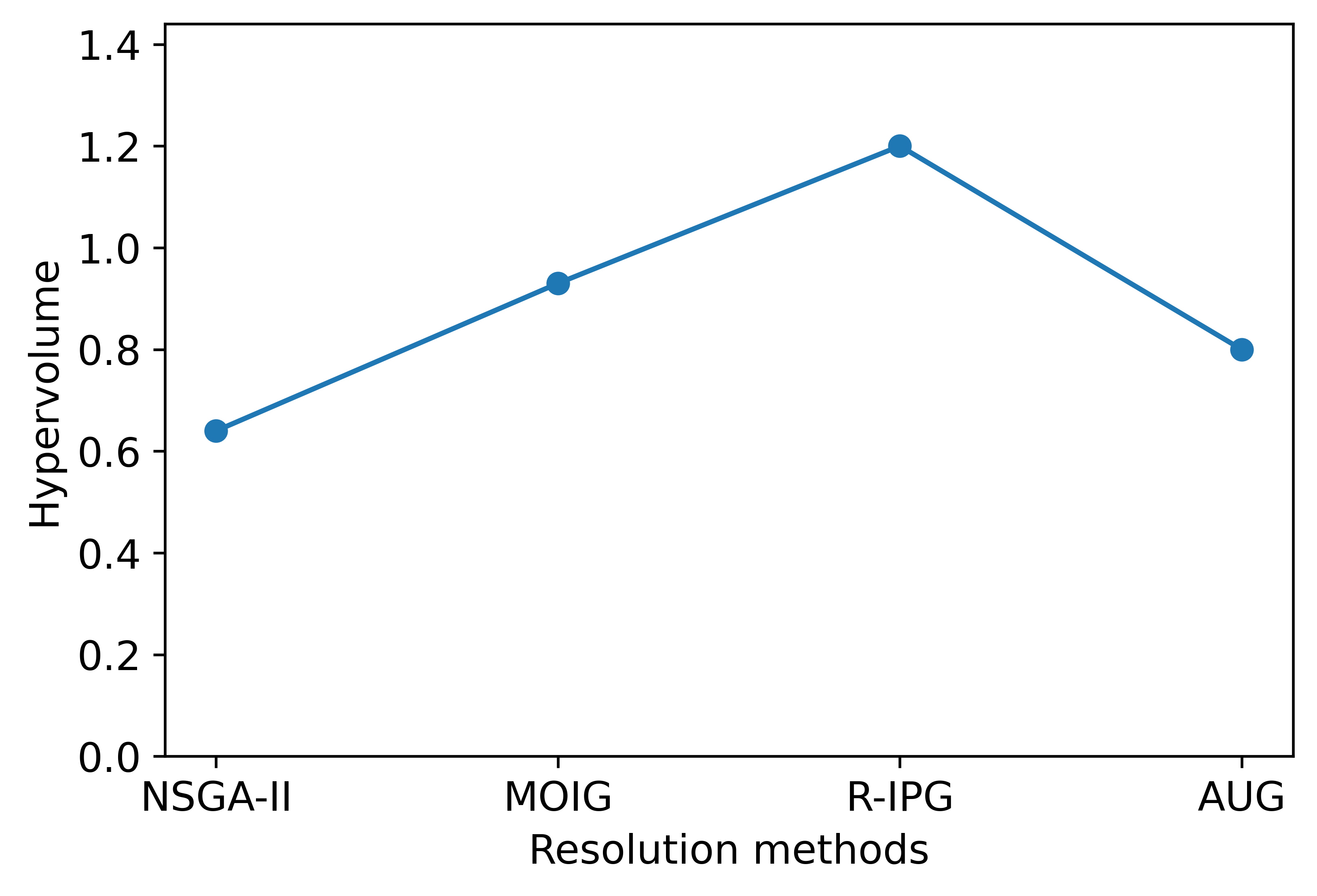}
        \label{fig:figure1}
    \end{minipage}%
    \begin{minipage}{0.5\textwidth}
        \centering
        \includegraphics[width=\linewidth]{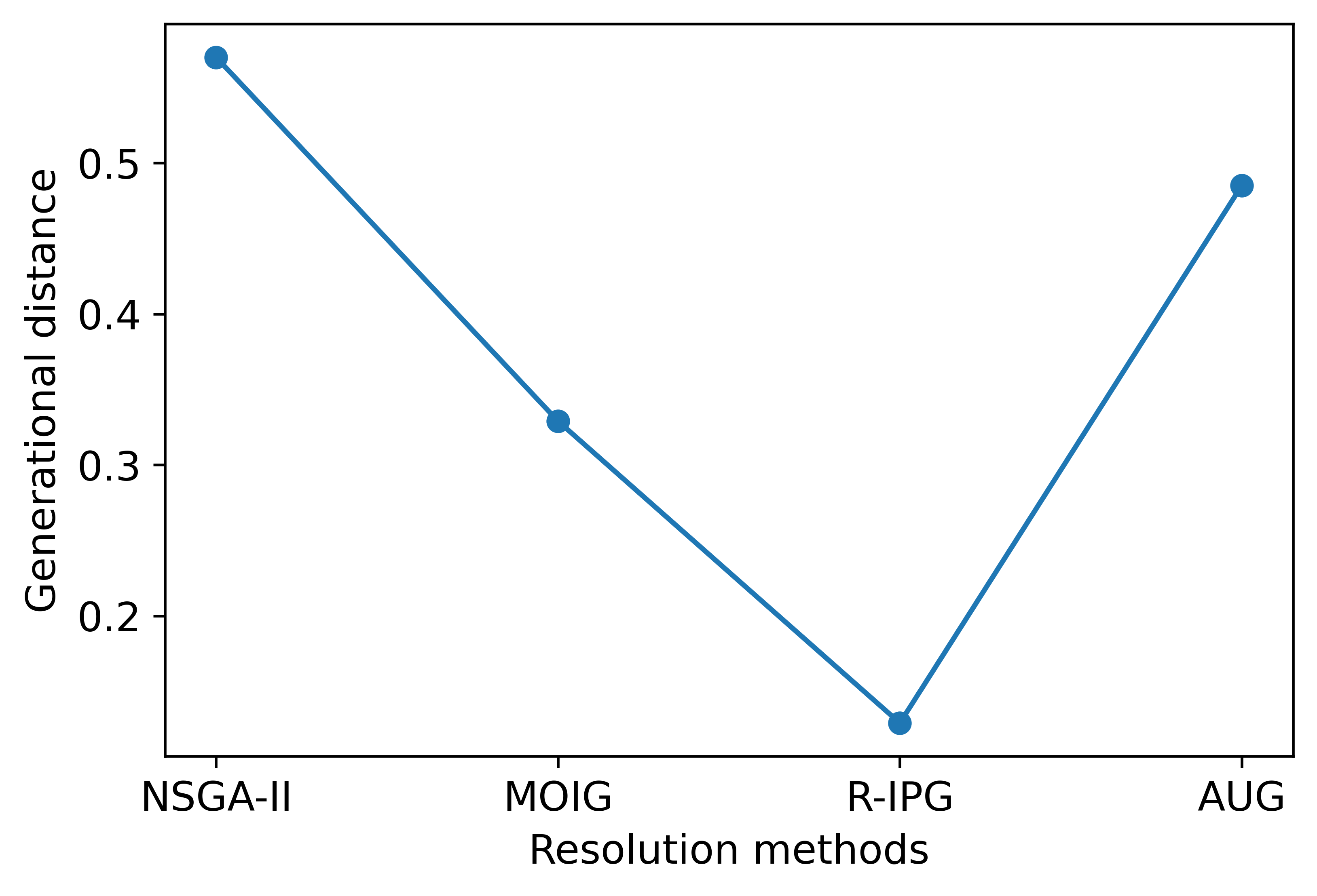}
        \label{fig:figure23}
    \end{minipage}
    \caption{$I_h$ and $GD$ means plots for all tested methods with medium instances}
    \label{means_plots_M}
\end{figure}

Figure~\ref{means_plots_M} presents the mean distribution of the ($I_h$) and GD across all medium-sized instances. The results clearly indicate that the R-IPG algorithm outperforms the other methods in terms of both convergence quality and solution diversity. Specifically, R-IPG consistently achieves higher $I_h$ values, reflecting a broader and more evenly distributed Pareto front, while also maintaining lower GD values, which indicate a closer approximation to the reference front.

Table \ref{Results_Large} shows the outcomes for the large instances. Following the results observed in the medium benchmark, R-IPG again appears as the top performer with an impressive average of \(I_h = 1.22\), while the AUG method provides less favorable results. 

\begin{figure}
    \begin{minipage}{0.5\textwidth}
        \centering
        \includegraphics[width=\linewidth]{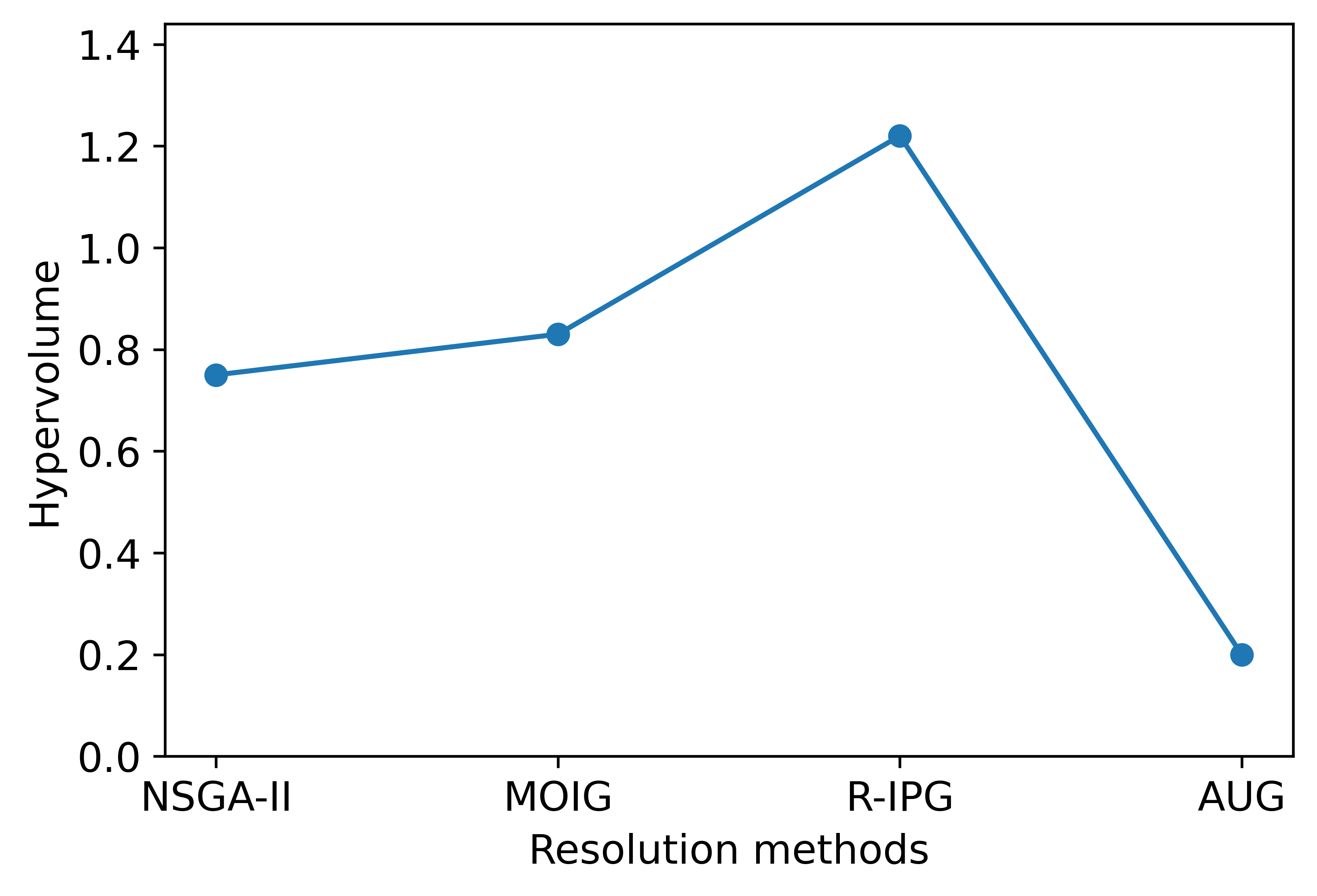}
        \label{fig:figure11}
    \end{minipage}%
    \begin{minipage}{0.5\textwidth}
        \centering
        \includegraphics[width=\linewidth]{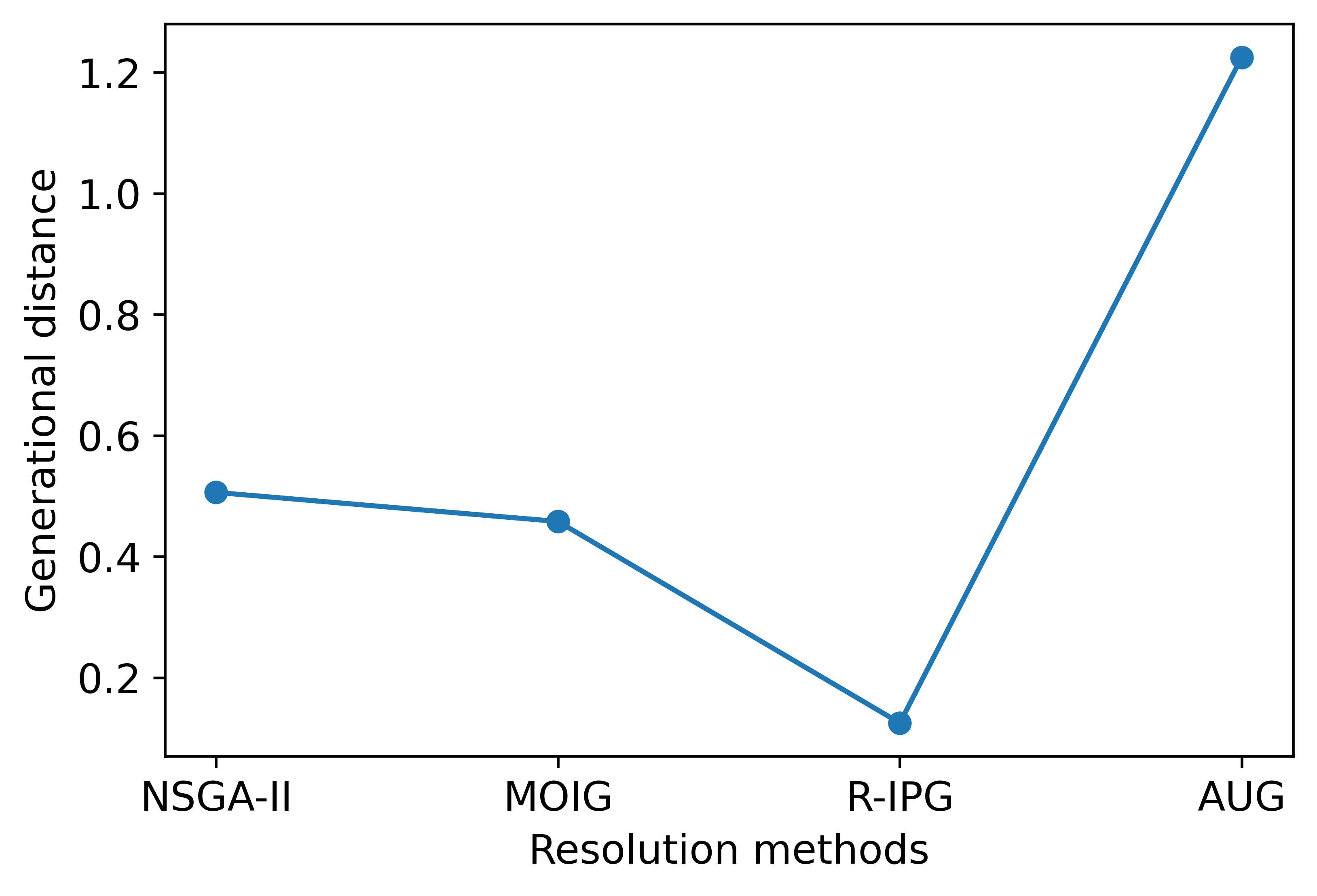}
        \label{fig:figure20}
    \end{minipage}
    \caption{$I_h$ and $GD$ mean plots for all tested methods with large instances}
    \label{means_plots_L}
\end{figure}

Figure~\ref{means_plots_L} demonstrates that the MOIG algorithm consistently outperforms NSGA-II across large-sized instances, positioning it as the second most effective method after R-IPG.

In summary, the experimental results demonstrate that the proposed R-IPG algorithm consistently delivers strong performance across all instance sizes. While the AUG method performs best for small problems, its effectiveness declines as instance size increases. In contrast, R-IPG maintains high-quality results, showing superior convergence and solution diversity for both medium and large instances. These findings confirm R-IPG as the most robust and scalable approach among the tested methods.


\begin{table}[t]

\caption{$I_h$ and $GD$ values for large instances (Best values are in bold)}

\label{Results_Large}
\tiny
\centering
\begin{tabular}{ccccccccccc}
 &  \multicolumn{3}{c}{}& \multicolumn{2}{c}{$I_h$} & \multicolumn{5}{c}{GD} \\
\hline
n	&	g	&	m	&	NSGA	&	MOIG	&	 RIPG	&	Aug	&	NSGA	&	MOIG	&	R-IPG	&	Aug	\\
\hline
50	&	2	&	2	&	0.51	&	0.91	&	\textbf{1.27}	&	0.04	&	0.629	&	0.306	&	\textbf{0.055}	&	1.335	\\
	&		&	3	&	0.92	&	1.18	&	\textbf{1.32}	&	0.08	&	0.372	&	0.223	&	\textbf{0.089}	&	1.305	\\
	&	3	&	2	&	1.27	&	1.34	&	\textbf{1.40}	&	0.47	&	0.099	&	0.054	&	\textbf{0.016}	&	0.958	\\
	&		&	3	&	0.42	&	0.29	&	\textbf{1.04}	&	N/A	&	0.689	&	0.776	&	\textbf{0.236}	&	N/A	\\
	&	4	&	2	&	1.21	&	1.24	&	\textbf{1.40}	&	0.28	&	0.600	&	0.788	&	\textbf{0.108}	&	N/A	\\
	&		&	3	&	0.56	&	0.43	&	\textbf{1.20}	&	N/A	&	0.162	&	0.152	&	\textbf{0.028}	&	1.147	\\
Avg	&		&		&	0.81	&	0.90	&	\textbf{1.27}	&	0.22	&	0.425	&	0.383	&	\textbf{0.089}	&	1.187	\\
\hline
100	&	2	&	2	&	1.28	&	1.37	&	\textbf{1.43}	&	0.37	&	0.101	&	0.044	&	\textbf{0.011}	&	1.092	\\
	&		&	3	&	0.34	&	0.25	&	\textbf{0.76}	&	N/A	&	0.811	&	0.872	&	\textbf{0.441}	&	N/A	\\
	&	3	&	2	&	1.07	&	1.27	&	\textbf{1.40}	&	0.11	&	0.953	&	0.967	&	\textbf{0.090}	&	N/A	\\
	&		&	3	&	0.25	&	0.23	&	\textbf{1.17}	&	N/A	&	0.279	&	0.128	&	\textbf{0.032}	&	1.327	\\
	&	4	&	2	&	0.94	&	1.09	&	\textbf{1.36}	&	0.04	&	0.383	&	0.268	&	\textbf{0.059}	&	1.414	\\
	&		&	3	&	0.22	&	0.30	&	\textbf{0.94}	&	N/A	&	0.999	&	0.919	&	\textbf{0.337}	&	N/A	\\
Avg	&		&		&	0.68	&	0.75	&	\textbf{1.18}	&	0.17	&	0.588	&	0.533	&	\textbf{0.162}	&	1.277	\\
\hline
G. avg	&		&		&	0.75	&	0.83	&	\textbf{1.22}	&	0.20	&	0.506	&	0.458	&	\textbf{0.125}	&	1.225	\\
\hline
\end{tabular}
\end{table}

\section{Conclusion and Future Work} \label{Conclusion}

In this paper, we introduced a multi-objective algorithm called Refined Iterated Greedy Pareto 
and a MILP model to address the multi-objective blocking hybrid flowshop with Makespan and TEC.
however, the structure of most industrial companies today can be considered 

This problem represents a generalized framework that includes many practical scenarios that exist in real-world production environments.
To solve this multi-objective scheduling problem, we used the augmented epsilon constraints, our proposed algorithm and two methods well selected from the literature. 
To compare the performances of different approaches, a benchmark of small, medium, and large instances is generated. The AUG method was only able to give the best Pareto fronts for small instances, however, its performance decreased when applied to medium and large benchmark cases. The proposed metaheuristic shows effectiveness in all instances.

Presented work can be extended by both problem and solution methodology. From the problem dimension, some other practical extensions such as as time of use, peak power, sequence-dependent setup times, and transportation times as well as distributed manufacturing can be added. The stochastic version of the problem can be considered to cover uncertainty originating from machine availability and processing times and simheuristic method \cite{juan2015review} can be developed which still use the heuristic proposed in this paper coupled with a discrete event simulation of BFHS for modeling the stochasticity. In addition to the mentioned uncertainty, the problem can be further extended to dynamic scheduling cases in which graph neural network based methods show promising results in other manufacturing systems \cite{JMStwo}.

From a solution methodology point of view, the current work can be extended in two ways as well. The first; developed heuristic algorithm can be implemented to other types of manufacturing settings different than BHFS as it brings promising results. The second, developed heuristic can be coupled with the MILP model developed in a large neighborhood search \cite{pisinger2019large}
setting in which while the heuristic is used for a good initial solution quickly, MILP model can be used for repairing destroyed solutions with the hope of improving the current best solution. In this way, the benefits of the heuristic and the exact methods; speed and quality can be fairly exploited even for the large instances. \\

\medskip \noindent \textbf{Acknowledgment.}
This publication has emanated from research conducted
with the financial support of Science Foundation Ireland under Grant number
12/RC/2289-P2 at Insight the SFI Research Centre for Data Analytics at
UCC, which is co-funded under the European Regional Development Fund and Grant number 22/NCF/DR/11264, National Challenge Fund, Digital for Resilience Challenge.
For the purpose of Open Access, the author has applied a CC BY public copyright licence to any Author Accepted Manuscript version arising from this submission.

\bibliographystyle{unsrtnat}
\bibliography{references}

\end{document}